# Sky detection and log illumination refinement for PDE-based hazy image contrast enhancement

Uche A. Nnolim[1]

*Department of Electronic Engineering, University of Nigeria, Nsukka, Enugu, Nigeria*

**Abstract**

*This report presents the results of a sky detection technique used to improve the performance of a previously developed partial differential equation (PDE)-based hazy image enhancement algorithm. Additionally, a proposed alternative method utilizes a function for log illumination refinement to improve de-hazing results while avoiding over-enhancement of sky or homogeneous regions. The algorithms were tested with several benchmark and calibration images and compared with several standard algorithms from the literature. Results indicate that the algorithms yield mostly consistent results and surpasses several of the other algorithms in terms of colour and contrast enhancement in addition to improved edge visibility.*

*Keywords:* fuzzy logic-based clustering; gradient and fuzzy edge detection; thresholding/segmentation and binary area computation; log illumination refinement; partial differential equations

1. Introduction

Images scenes acquired in poor weather are affected by scattering and light absorption by particles suspended in the atmosphere and exhibit hazy appearance [1]. These images with poor contrast and visibility, need to be restored by de-hazing techniques, which can be based on multiple or single image methods. These can be achieved by physical hardware/optical or digital or software-based medium and perform visibility restoration revealing details in the image scenes [1]. Software-based single image de-hazing algorithms are cheaper, faster and easier to devise and implement and are preferable to multi-image de-hazing and physical, optics-based methods. These latter schemes are unfeasible, expensive and time-consuming for most existing hazy image scenes due to unavailability of image data under varying conditions and factors. Thus, the proliferation of single image de-hazing algorithms is unsurprising and constitute a sizeable portion of the image de-hazing literature [1] [2] [3] [4] [5] [6] [7] [8] [9] [10] [11] [12] [13] [14] [15] [16] [17] [18] [19].

Single image de-hazing methods are usually grouped under restoration- or enhancement-based frameworks [1]. The enhancement methods are based on application of image contrast enhancement algorithms from a variety of domains to solve the de-hazing problem. Examples of algorithms in this group include contrast limited adaptive histogram equalization (CLAHE) [20], Retinex [8] [13] [16] in addition to variational contrast-based de-hazing [21] and recent partial differential equation (PDE)-based formulations [22]. However, these methods by default, generally result in colour distortion [13] or over-enhancement of hazy images, grey-world violation and halo effect in Retinex-based methods. This is in addition to over-exposure or over-enhancement of sky regions and under-exposure of dark regions, resulting in dark images.

The restoration-based approach is based on the physical hazy image formation model, which results in the recovery of the radiance map as the original, haze-free image [1]. However, due to the fact that de-hazing process is an ill-posed, under-constrained problem, several attempts to model the degradation prior (from which the radiance image is recovered), have been proposed [1]. The key works that have arguably made the restoration-based approach more popular are those by Tan [1], Fattal [2] [3], and the dark channel prior (DCP) method by He, et al [4]. The DCP-based methods exhibit visible halo effects and darkening of image results. Several solutions have been proposed to address this problem [23] [17] [5] [24] [23] [25] [7] [26] [11] with mixed results as most algorithms aim to improve the results based on pre- or post-processing combined with existing (or modifications of the) DCP-based framework [10]. Some examples include improved dark channel prior (IDCP) [17], improved atmospheric scattering models [27], regression model [28], Graph cut method [29] and adaptive bi-channel priors on super pixels [30]. Other works include those based on sky segmented DCP [31], large sky region segmentation and multiscale opening dark channel model [32], and segmentation [7] [26] [33], fusion [34] [35], geometry [36], Weighted Least Squares [37] and high dynamic range (HDR) tone mapping [38]. Additional works are those utilizing variational- [39] [40] [35] [41], regularization- [26] [42] [43], retina- [44] multi-scale convolutional neural networks- [45] based methods. Other problems introduced by several of these algorithms range from increased computational complexity to manual tuning of numerous parameters, rendering them impractical for batch processing of diverse, hazy images. Furthermore, several perform minimal enhancement of hazy images.

A previous PDE-based formulation for hazy image enhancement was presented in previous work and yielded good results [22]. However, it also introduced additional issues, which were addressed. One of the key issues was the image darkening and over-enhancement problem. This was addressed by determining which of either the illumination or reflectance channel to process. However, there was no automated method for determination of which component required enhancement and results were manually determined. In the current work, automated sky detection and enhancement and log illumination refinement schemes were proposed [46] and extensive experiments utilizing several benchmark images indicated improvements in the results over the previous proposed scheme. Additionally, image and quantitative metric-based comparisons with several algorithms from the literature showed that the scheme performed more consistently and better than several works from the literature. Moreover, the results are presented in this work to provide extensive evidence of the claims of the proposed approach.

The rest of the paper is as follows; the second section provides relevant background on image de-hazing algorithms. The third section presents the proposed modified algorithm and novel key features. The fourth section presents the results of experiments and comparisons with established works from the literature. The final section presents the conclusions.

## 2. Prior related work

The segmentation-based schemes were developed to solve the problem of sky region over-enhancement as this is caused by the non-uniformity of haze across sky and non-sky regions [5]. The improved DCP (IDCP) by Xu, et al [17] and a soft threshold proposed by Liu, et al [11] attempted to solve the problem by using a hard threshold. This was followed by the adaptive sky detection and preservation scheme for de-hazing by Dai et al [5] and a haze density aware de-hazing scheme using adaptive parameter based on exposure and YCbCr colour saturation adjustment by [9]. Subsequently, Mao, et al [25] proposed a haze degree detection and estimation method for hazy images. The sky detection and segmentation-based schemes help improve DCP results by reducing halos, etc. However, several of these schemes can also result in minimal contrast enhancement, colour distortion, under/over-enhancement of sky and detail regions, yielding darkening and discolouration in de-hazed images. For example, the method by Anwar, et al (WLS and high dynamic range (HDR) tonal mapping post processing method) [38], causes colour distortion and greying out of sky regions, while the method by Ju, et al (improved atmospheric scattering model) [27] results in discolouration in addition to halo effects with considerable haze still present in some images and sky region over-enhancement. Additionally, some of these algorithms are highly complex and require the selection and tuning of several parameters. For example, the work by Liu, et al (large sky region segmentation and multiscale opening DC model) [32] requires training of a two-class classifier to determine sky/non-sky region pixels and also results in discolouration/artificial colours for sky region while haze is not always adequately removed. Furthermore, the method by Gu, et al (Average Saturation Prior (ASP) combined with improved atmospheric scattering models and guided total variation model) [47] requires that key parameters are obtained via training with vast amounts of hazy image data [47]. In addition, performance depends on the clustering number, $k$, based on the $k$-means algorithm utilized, which requires that results are to be refined using Markov Random Field (MRF) model [47].

## 3. Proposed algorithms

The proposed algorithms are the sky detection and enhancement scheme (SDS) followed by the log illumination refinement scheme (LIR/LIRS) [46]. Both are meant to be utilized by the previously proposed PDE-based formulation of Retinex and CLAHE combination (PA-2), which allows the use of the illumination/reflectance (I-R) model to enhance images [22]. The SDS is the more complex of the two and is composed of three key stages namely; gradient/homogeneity map evaluation, thresholding and binary region/area computation and homogeneity ratio calculation. This ratio value is utilized in a decision stage-based version of PA-2. The diagram for the proposed sky detection algorithm is shown in Fig. 1.

### 3.1 Sky detection and enhancement (SDS)

The gradient/homogeneity image depicts the changes in the image scene composed of homogeneous (smooth) regions and non-homogeneous (detail) regions. These are normally characterized as low frequency and high frequency components in the Fourier domain or illumination and reflectance components in the spatial domain using the I-R model. They can also be denoted as non-edge and edge-based regions in the gradient domain. Thus, the smooth regions will have mostly constant values, thus the image gradient is zero ($|\nabla I| = 0$) in these regions. Conversely, the detail (non-smooth) regions will have a non-zero gradient ($|\nabla I| > 0$). These two regions are best resolved in binary terms as black and white regions, where the dark areas are inactive/constant regions, while the light areas are the active regions/zones. Thus, computing the number of pixels in each region determines the area and its contribution to the entire image scene.

This is then followed by the binary thresholding/segmentation to simplify the required data needed for the decision stage. The binary image then allows us to compute the black and white regions to compute the homogeneity ratio. If the image has minimal or no sky regions, we expect the dark pixel area to be extremely miniscule, if not close to zero and conversely, the white pixel area should contribute almost the entire pixels in the image. Thus, the homogeneity ratio is the ratio of black pixel area to white pixel area and is nearly zero for images with minimal or no sky regions and greater than zero for images with considerable sky regions. This is presented in the classification-based verification of the proposed method using a specified hazy image dataset.

The edge detection and thresholding algorithms utilized are crucial to the final result and there are several possible methods to compute the gradient map and aid the thresholding operation. The gradient map can be obtained using linear spatial filter kernels, such as Sobel, Prewitt, etc [48]. or non-linear methods using statistical methods of dispersion in addition to fuzzy-rule-based edge detection schemes. Additionally, the thresholding can be performed using algorithms such Otsu [49], Entropy [50], ISODATA [51], fuzzy *c*-means clustering methods [51], etc. Thus, the mode and objective of preliminary experimentation here is to determine the best edge detection method and combine with the best thresholding algorithm all within the context of run-time and reliability. Subsequently, the system is implemented, verified for consistency and actually improves upon the previous algorithm.

In experiments, the non-linear filters were the most flexible, while the linear filters where the least flexible. However, this presented another problem; the need to determine the optimum window size of the non-linear filter that yields best results for all images without considerable increase in computational complexity. This would enable the avoidance of morphological region growing algorithms and reduce the execution time of the algorithm.

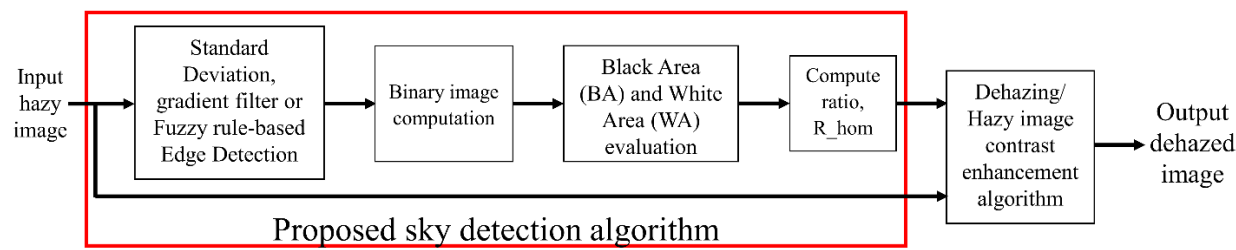

**Fig. 1** Automated sky region detection combined with hazy image contrast enhancement scheme

**3.1.1 Selection of edge detection algorithm**
In this section, we compare the known edge detection algorithms to obtain the best possible results in terms of edges and runtime. The edge filter algorithms tested include;
- Roberts, Sobel, Prewitt, Canny, Logarithm of Gaussian (LoG) and Zero-crossing filters [48].
- Statistical filters such as Standard Deviation (SDF), Median Absolute Deviation (MADF), average absolute deviation (AADF), Range (RF), Interquartile Range (IQRF) and Mean Difference (MDF) and Relative Mean Difference (RMDF) filters [48] and Fuzzy edge detection filters [52].

We focus on the statistical edge detectors and the comparison and plots of the runtimes of the statistical edge filters in relation to window size is shown in Fig. 2 and 3. The linear edge detectors were deselected due to lack of flexibility and the output is a linear combination of the inputs, thus increase in size increases noise around edges as seen in Fig. 4 and false edges are created. Thus, the linear detectors seem to work best in the absence of haze.

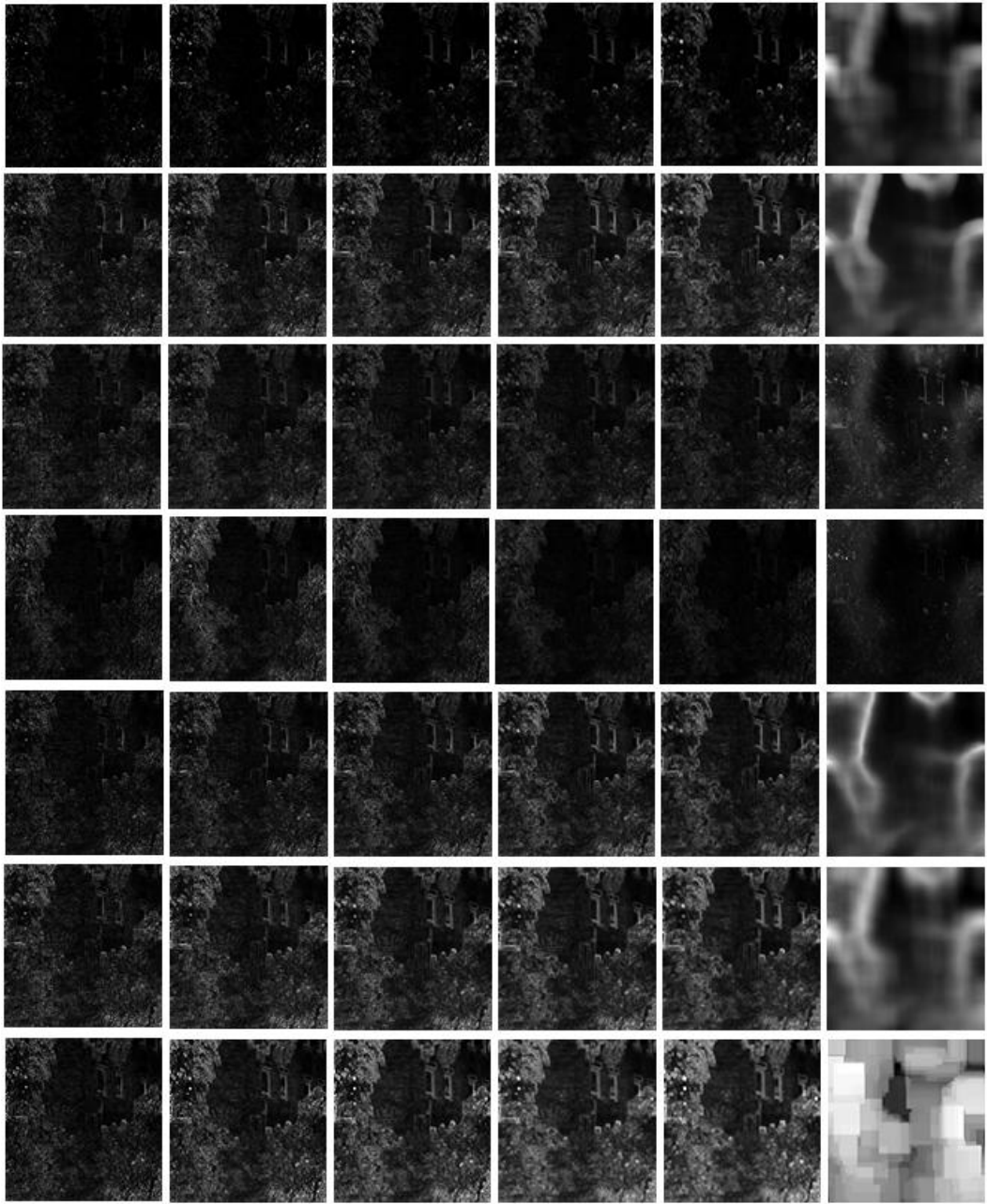

(a)

|       |     |     |     |     |       |         |
|-------|-----|-----|-----|-----|-------|---------|
| **SDF**   | 3×3 | 5×5 | 7×7 | 9×9 | 11×11 | 135×135 |
| **IQRF**  | 3×3 | 5×5 | 7×7 | 9×9 | 11×11 | 135×135 |
| **MDF**   | 3×3 | 5×5 | 7×7 | 9×9 | 11×11 | 135×135 |
| **RMDF**  | 3×3 | 5×5 | 7×7 | 9×9 | 11×11 | 135×135 |
| **MADF**  | 3×3 | 5×5 | 7×7 | 9×9 | 11×11 | 135×135 |
| **AADF**  | 3×3 | 5×5 | 7×7 | 9×9 | 11×11 | 135×135 |
| **RF**    | 3×3 | 5×5 | 7×7 | 9×9 | 11×11 | 135×135 |

(b)

**Fig**. **2** (a) Comparison of various statistical edge filters for varying window sizes (b) key to figures

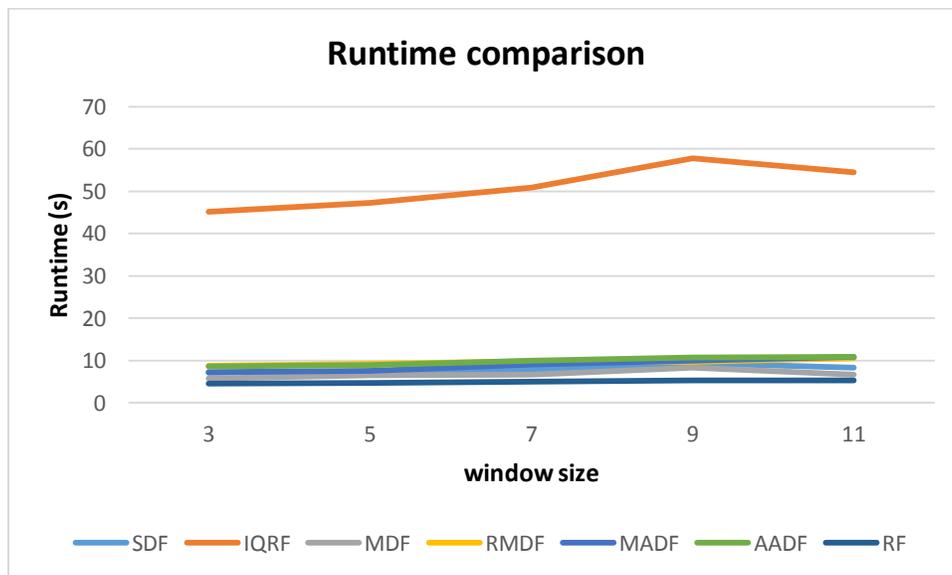

**Fig. 3** Runtime comparison of statistical edge detection filters for varying window sizes

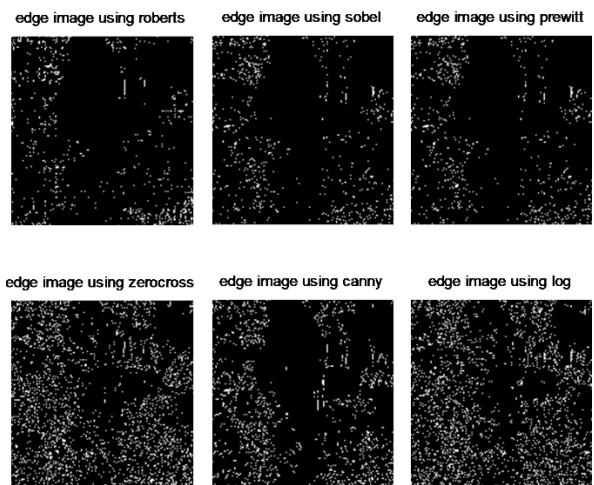

**Fig**. **4** Comparison of various linear edge filters

Based on the experiments, the linear edge detectors were the fastest with least complexity while the fuzzy edge detectors were the most complex but more effective at capturing larger regions when thresholded, thus eliminating the need for morphological region growing or shrinking operations. For the statistical operators, they were moderately complex and were adjustable in terms of window size, which enabled them to capture larger regions as well. However, increasing window size also increased computational complexity, raising the number of operations per pixel. Also, though the range filter consistently has the lowest runtime and highlights edges more, it introduces blocky effects for large window regions, unlike the others. Apart from the MDF and RMDF, the rest show good blending of edge regions, though AADF is blurrier while the IQRF is the best though its runtime is the highest. This leaves MADF and SDF, though the former requires much more runtime and operations than the SDF. Thus, we select the SDF as the best compromise. The fuzzy-based and linear edge detectors represent the extremes of best accuracy but highest complexity and least effective but fastest and lowest complexity. Additionally, the rule-based fuzzy edge detectors are still mostly based on the linear edge and some statistical edge detectors. We also explored various binary thresholding algorithms and chose Otsu's method [49] due to its performance and the fact that the edge detection scheme is key to obtaining an accurate homogeneity map, based on experiments.

### 3.2 Log illumination refinement
The logarithmic illumination refinement basically finds and updates pixels less than the maximum log illumination value with the maximum value and passes this on to the previously proposed algorithm (PA-2) [46]. Thus, this scheme is less computationally intensive than the SDS scheme but also less adaptive and can lead to colour fading in attempts at haze removal. This scheme is a more direct approach at hazy image illumination-based enhancement.

### 4. Experiments and results
We now present the results of experiments for the proposed addition to PA-2. The images utilized are taken from the FRIDA3 and FIDA data set by Tarel, et al [53] [54]. Additionally, popular benchmark images employed in numerous image de-hazing papers and other images from the internet were also utilized in experiments. Algorithms compared include those by Fattal [2], Tan [1], Kopf, et al [55], Fattal [3], He, et al [4], Tarel, et al [56], Galdran, et al [39], Zhang, et al [44], Oakley & Bu [57], Kim, et al [21], Dai, et al [5], Hsieh, et al [58], PA1 [22], PA2 [22], PA2 + SDS, PA2 + LIR, Zhu, et al [59], Ren, et al [45], Anwar, et al [38], Liu, et al [32], Ju, et al [27], etc. The measures utilized for the experiments include the (Relative) Average Gradient [60] (RAG/AG), hue deviation index (HDI) [60], colourfulness (C) and color enhancement factor (CEF) [61], ratio of visible edges, $Q_e$, (ROVE), ratio of black and white areas, σ (RBWA) in addition to mean structural similarity index (MSSIM) [62], Peak Signal to Noise Ratio (PSNR), Mean Absolute Error (MAE) and Mean Square Error (MSE). We also verify the algorithms using fuzzy $c$-means clustering to classify the obtained ratios and evaluate accuracy of the measure. Results are presented accordingly.

### 4.1 FRIDA3 dataset
This dataset consists of 66 synthetic images each of both left and right views with and without fog, resulting in about 264 images. It can be used to calibrate and benchmark de-hazing algorithms. Additionally, because the reference images exist, full-reference metrics such as SSIM, PSNR are applicable here. Fig. 5 shows the classification results using fuzzy c-means clustering algorithm to verify the validity of the SDS method. As expected, all values greater than zero are classified as having some sky region, which is true of all the images in this dataset. Fig. 6 shows the original images, their hazy versions and results obtained from processing with algorithms by Zhu, et al, He, et al, PA-2, PA-2+SDS and PA-2+LIRS. Results show the discolouration effects of PA-2 are evident (purple sky region). This effect is resolved by the SDS and LIRS schemes. The method by He, et al and Zhu, et al show dark and faded images respectively with fewer enhanced details compared to PA-2, PA2+SDS and PA2+LIRS. The average values of the metrics are presented in Table 1 and show that the PA2+SDS method yields the best results for most of the measures utilized and the quantitative metrics appear to corroborate the visual results.

### 4.2 FIDA dataset and benchmark images
The FIDA dataset consists of 13 images while the benchmark images include those normally used in de-hazing experiments (in addition to images from the internet) to test the efficacy of the algorithm. The designation of a sample of the image homogeneity maps and ratio computation are presented in Fig. 7 and compared with the actual images. The classification results are also presented in Fig.8 using fuzzy $c$-means clustering algorithm [51] to compare assumptions. In this case, some images do not have sky regions and have a ratio of zero, as expected. The images obtained from the internet all have sky regions since they are aerial views of cities with skyline views. We also present results of the algorithms against several other algorithms from the literature in Fig. 9. Results show that the proposed approaches help improve the results of PA-2 and also surpass the results of several other algorithms as shown.

We also compare results against methods by Fattal, He, Kopf, Tan, Tarel and Zhu for the colour images in the FIDA dataset in Fig. 10. The most colourful results are from PA-2+SDS, PA2+LIRS, Tan and PA2, though PA2 shows over-enhancement of colour, while Tan doesn't yield as much detail (mountain region of last image) compared to the proposed schemes.

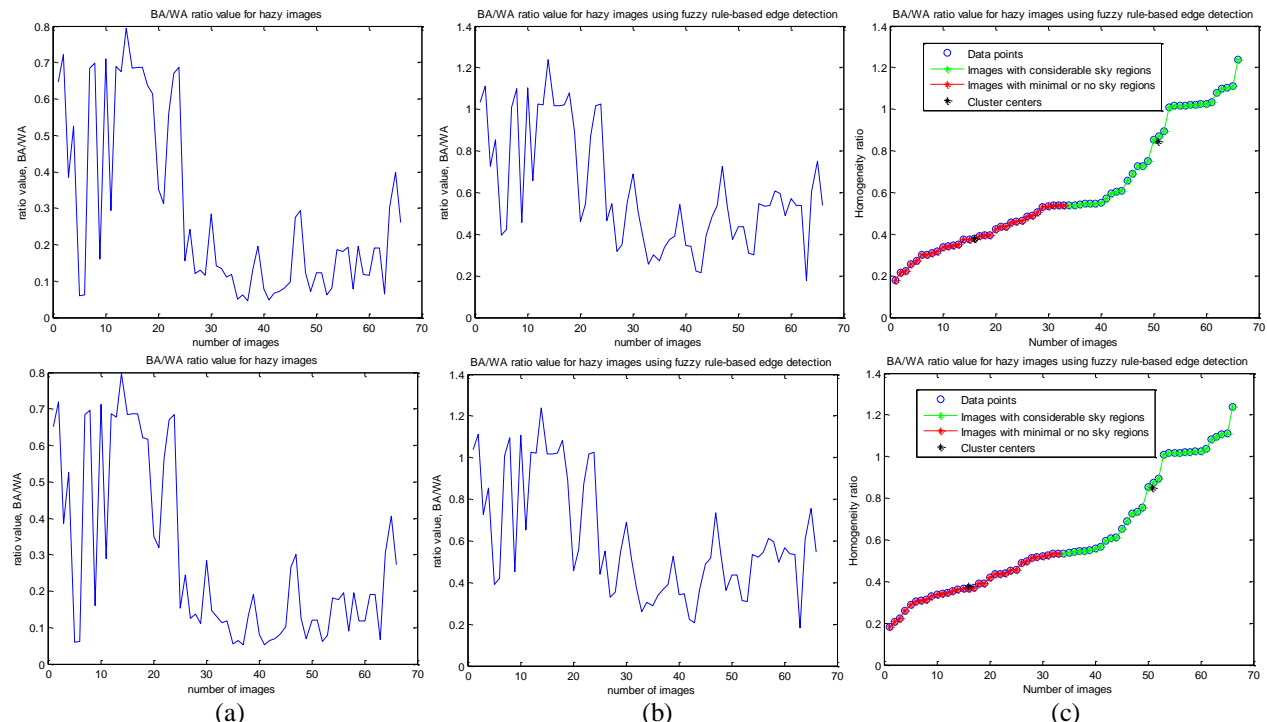

**Fig. 5** Plot of ratio values for 66 Left view (first row) and Right view (second row) hazy images from FRIDA3 dataset (third row) using (a) standard deviation filter with large kernel (b) fuzzy-rule-based-edge detection (c) grouping of sorted image ratio values using fuzzy clustering method

**Table 1** Average values for 66 (a) left and (b) right view (FRIDA database) images processed with Zhu, et al (CAP), He, et al, (DCP), PA-2, PA-2 + SDS and PA-2 + LIR

| Algorithm | Qe | BWAR | RAG | MSSIM | PSNR | MAE | MSE |
|---|---|---|---|---|---|---|---|
| CAP | 0.967408 | 0.01001 | 1.617735 | 0.592498 | 11.09096 | 53.86885 | 5415.557 |
| DCP | 0.97796 | 0.00017 | 2.19491 | 0.648064 | 12.56377 | 44.63618 | **3848.073** |
| PA2 | 1.058611 | 8.8E-05 | 2.276011 | 0.611011 | 11.00526 | 63.69751 | 5464.016 |
| PA-2 + SDS | **1.254145** | 1.59308E-05 | **2.632071** | **0.71882** | **12.67592** | **43.88135** | 3910.314 |
| PA-2 + LIR | 1.115049 | **5.86924E-05** | 2.445847 | 0.693955 | 12.59826 | 45.75688 | 3901.96 |

(a)

| Algorithm | Qe | BWAR | RAG | MSSIM | PSNR | MAE | MSE |
|---|---|---|---|---|---|---|---|
| CAP | 0.933719 | 0.00972 | 1.619955 | 0.593161 | 11.0824 | 53.90755 | 5422.409 |
| DCP | 0.954648 | 0.00017 | 2.19394 | 0.648356 | 12.54855 | 44.71968 | **3856.819** |
| PA2 | 1.03751 | 1.13439E-06 | 2.2759 | 0.609253 | 10.95889 | 64.02167 | 5537.211 |
| PA-2 + SDS | **1.214761** | 1.16892E-05 | **2.634608** | **0.71873** | **12.66522** | **43.97349** | 3917.517 |
| PA-2 + LIR | 1.124608 | 1.02588E-05 | 2.447449 | 0.693724 | 12.58288 | 45.87162 | 3914.086 |

(b)

| Img no. | Image name | Image | Designation & Exaggerated Gradient map | Ratio value & binary map |
|---|---|---|---|---|
| 1 | '9.png' | 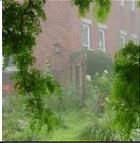 | No Sky 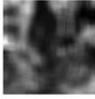 | 0/0.0035 |
| 2 | 'canon.jpg' | 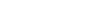 | No Sky 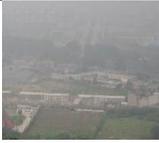 | 0/0.0678 |
| 3 | 'canon3.bmp' | 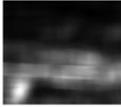 | Sky 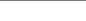 | 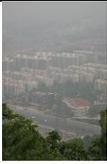 0.0699/0.1433 |
| 4 | 'canyon.bmp' | 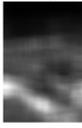 | No Sky 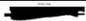 | 0/ 0.0445 |
| 5 | 'city_1.jpg' | 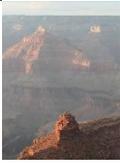 | Sky 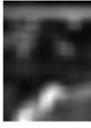 | 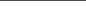 0.0971/0.2454 |
| 6 | 'city_2.jpg' | 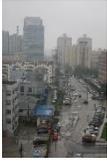 | Sky 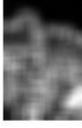 | 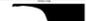 0.059/ 0.2613 |
| 7 | 'cones.jpg' | 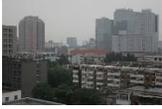 | No Sky 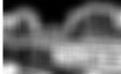 | 0/ 0.0028 |
| 8 | 'forest.jpg' | 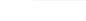 | No Sky 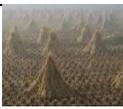 | 0/ 0.0008 |
| 9 | 'gugong.bmp' | 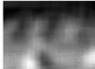 | No Sky 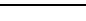 | 0/ 0.005 |
| 10 | 'hongkong.bmp' | 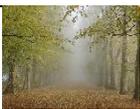 | Sky 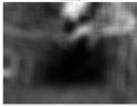 | 0/ 0.0583 |

| 11 | 'lake.jpg' | 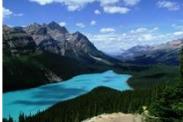 | Sky | 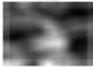 | 0/ 0.0171 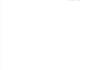 |
| --- | --- | --- | --- | --- | --- |
| 12 | 'mountain.jpg' | 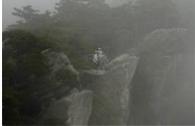 | No Sky | 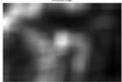 | 0/ 0.0407 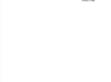 |
| 13 | 'ny1.bmp' | 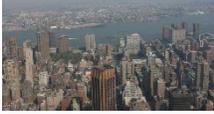 | Sky | 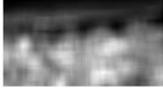 | 0/ 0.0408 |
| 14 | 'ny_1.jpg' | 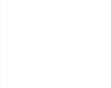 | Sky | 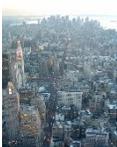 | 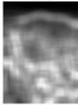 0.0565/ 0.1291 |
| 15 | 'ny_2.jpg' | 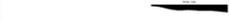 | Sky | 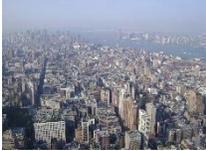 | 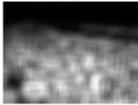 0.001/ 0.0691 |
| 16 | 'pumpkins.jpg' | 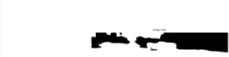 | No Sky | 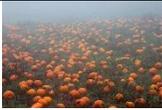 | 0/ 0.0032 |
| 17 | 'stadium.jpg' | 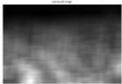 | No Sky | 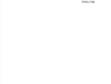 | 0/ 0.0009 |
| 18 | 'tiananmen1.png' | 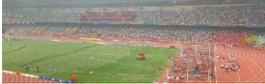 | Sky | 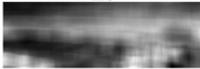 | 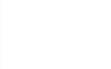 0.1826/ 0.4918 |
| 19 | 'toys.jpg' | 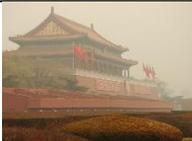 | No Sky | 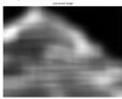 | 0/ 0.0669 |
| 20 | 'train.bmp' | 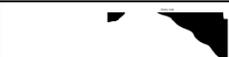 | No Sky | 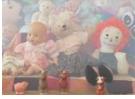 | 0/ 0.0094 |

**Fig. 7** Original hazy images and their exaggerated gradient map and designation along with binary map and ratio values

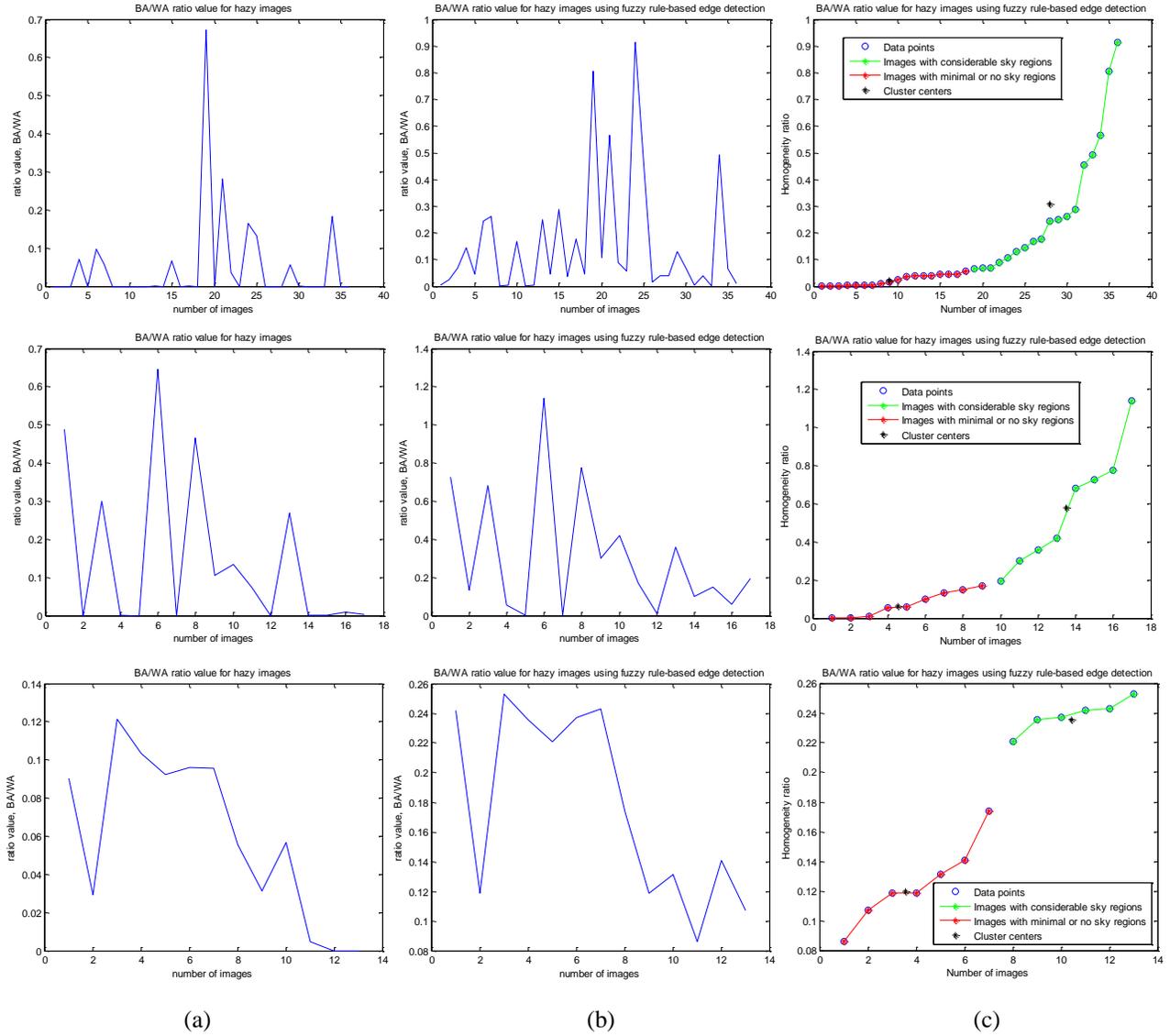

(a) (b) (c)

**Fig. 8** Plot of ratio values for 36 benchmark (first row), 17 internet (second row) and 13 hazy images from FIDA dataset (third row) using (a) standard deviation filter with large kernel (b) fuzzy-rule-based-edge detection (c) grouping of sorted image ratio values using fuzzy clustering method

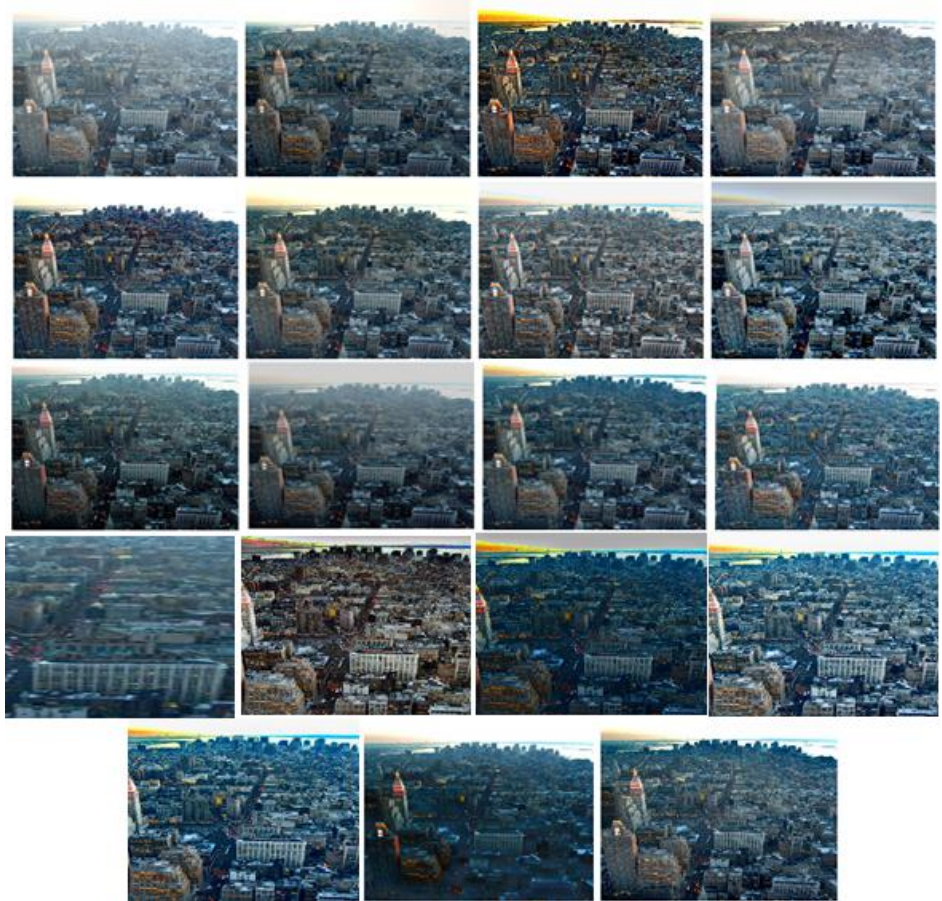

**Fig. 9**-(1) Image results and (2) key to figures: (a) Original hazy image (b) Fattal [2] (c) Tan [1] (d) Kopf, et al [55] (e) Fattal [3] (f) He, et al [4] (g) Tarel, et al [56] (h) Galdran, et al [39] (i) Zhang, et al [44] (j) Oakley & Bu [57] (k) Kim, et al [21] (l) Dai, et al [5] (m) Hsieh, et al [58] (n) PA1 [22] (o) PA2 [22] (p) PA2 + SDS (q) PA2 + LIR (r) Zhu, et al [59] (s) Ren, et al [45]

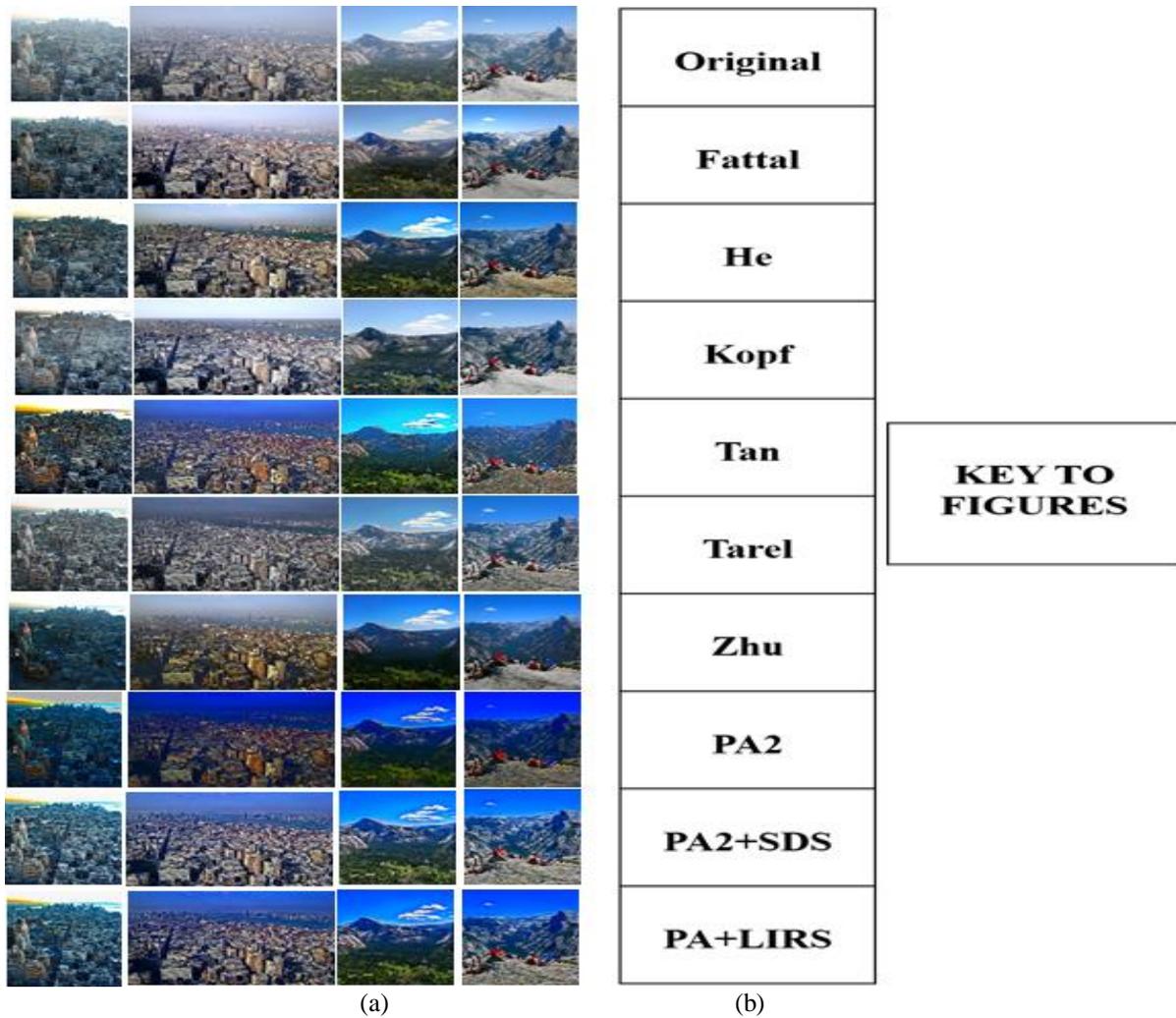

(a)          (b)

**Fig. 10** Original hazy images processed with algorithms by Fattal [2] [3], He, et al [4], Kopf, et al [55], Tan [1], Tarel, et al [56], Zhu, et al [59], PA2 [22], PA2 + SDS and PA2 + LIR

We also compare the results of processed greyscale images from the FIDA dataset with the same algorithms from Fig. 10 in Fig. 11 and the proposed schemes yield the some of the best contrast enhancement in addition to Tarel's method. However, PA-2 yields over-enhanced sky region, which is avoided by PA2+SDS and PA2+LIRS.

We compare the proposed improvements with various algorithms from the literature for the popular benchmark images in Fig. 12 and observe that they yield mostly consistent results compared with the other methods. We also compare the execution times of the proposed schemes with methods by He, et al, Zhu, et al and Ren, et al in Fig. 13 and though the algorithms are not as fast as the others, they are quite effective based on the amount of improvements observed in their image results.

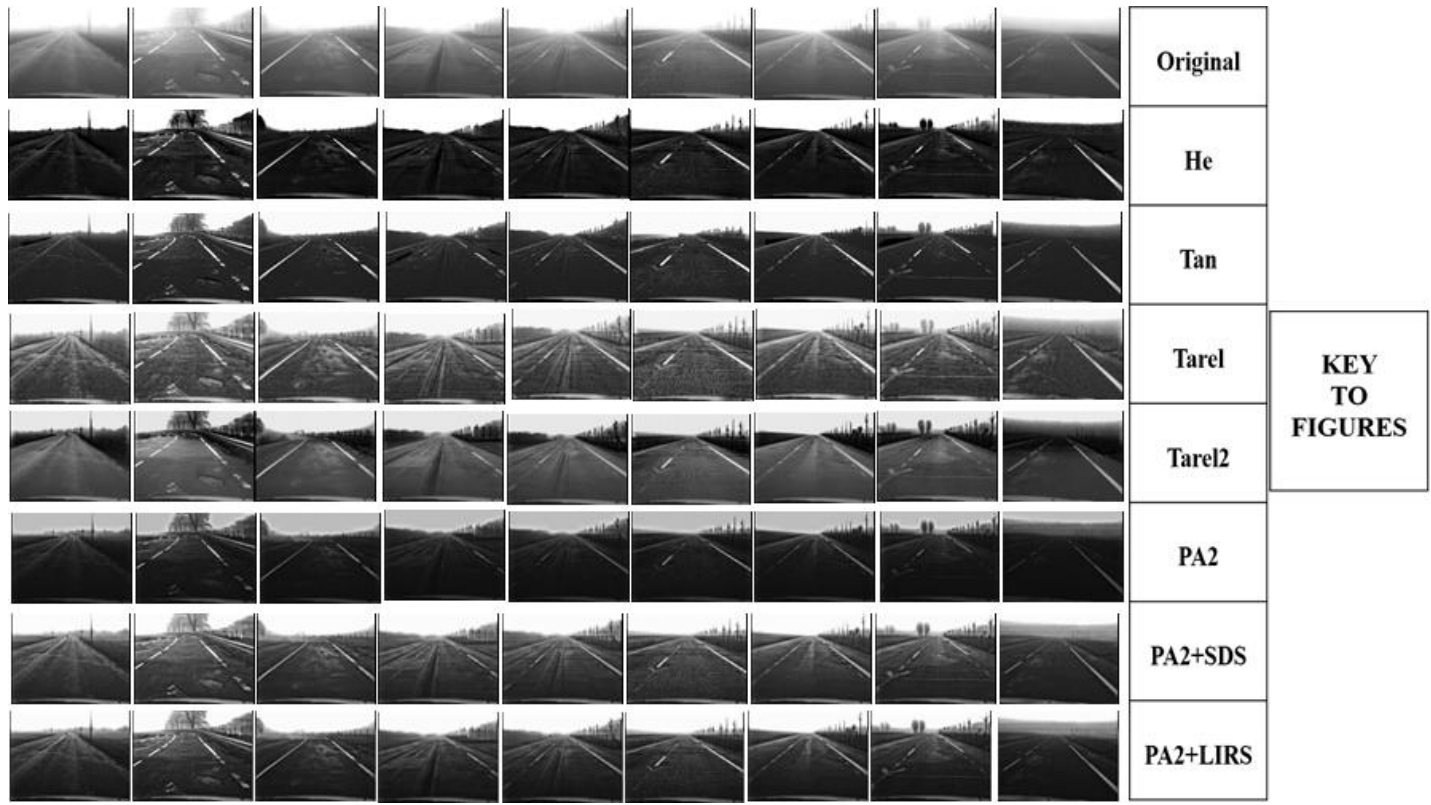

**Fig. 11** Original hazy images processed with algorithms by He, et al [4], Tan [1], Tarel, et al [56], PA2 [22], PA2 + SDS and PA2 + LIR

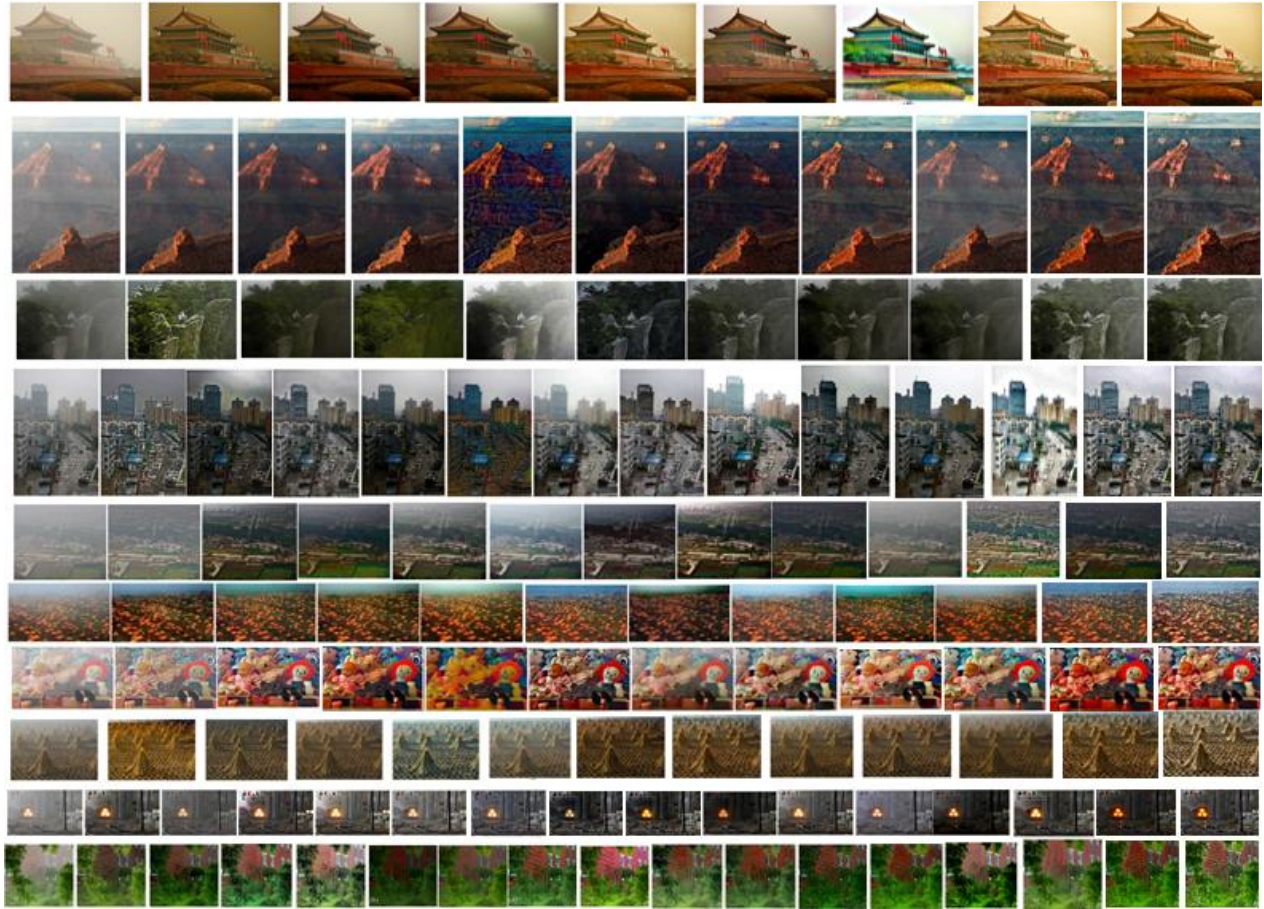

(a)

| Original | Tarel & Hautiere | Zhu, et al | He, et al | PA-2 | Ren, et al | Ju, et al | PA-2 + SDS | PA-2 + LIRS | | | | | | | Tiananmen image |
|---|---|---|---|---|---|---|---|---|---|---|---|---|---|---|---|
| Original | Fattal | Tan, et al | Yang, et al | Nishino, et al | Zhu, et al | He, et al | PA-2 | Ren, et al | PA-2 + SDS | PA-2 + LIRS | | | | | Canyon image |
| Original | Tarel & Hautiere | He, et al | Nishino, et al | Galdran, et al (EVID) | Wang & He | PA-2 | Ren, et al | Zhu, et al | PA-2 + SDS | PA-2 + LIRS | | | | | Mountain image |
| Original | Tarel & Hautiere | He, et al | PA-2 | Zhu, et al | Nishino, et al | Galdran, et al (EVID) | Wang & He | Dai, et al | Ren, et al | Liu, et al | Ju, et al | PA-2 + SDS | PA-2 + LIRS | | City1 image |
| Original | Tarel & Hautiere | He, et al | Nishino, et al | Ren, et al | Galdran, et al (EVID) | Wang & He | Dong, et al | PA-2 | Zhu, et al | Anwar, et al | PA-2 + SDS | PA-2 + LIRS | | | Canon image |
| Original | Fattal | Dong, et al | He, et al | Yeh, et al | PA-2 | Kratz & Nishino | Ancuti, et al | Ren, et al | Zhu, et al | PA-2 + SDS | PA-2 + LIRS | | | | Pumpkins image |
| Original | Tarel & Hautiere | Dai, et al | He, et al | Nishino, et al | PA-2 | Galdran, et al (EVID) | Wang & He | Zhu, et al | Ren, et al | PA-2 + SDS | PA-2 + LIRS | | | | Toys image |
| Original | Nishino, et al | Tarel & Hautiere | Meng, et al | Galdran, et al (EVID) | Galdran, et al2 (FVID) | He, et al | PA-2 | Hsieh, et al | Ren, et al | Zhu, et al | PA-2 + SDS | PA-2 + LIRS | | | Cones image |
| Original | He, et al | Tarel & Hautiere | Meng, et al | Gibson, et al | Galdran, et al (EVID) | Dai, et al | PA-1 | PA-2 | Fattal | Lu, et al | Ren, et al | Zhu, et al | Liu, et al | PA-2 + SDS | PA-2 + LIRS | Train image |
| Original | Hsieh, et al | Zhu, et al | Kim, et al | Ancuti, et al | He, et al | Tarel & Hautiere | Zhang, et al | Yeh, et al | Fattal | Dong, et al | Guo, et al | PA-2 | Ren, et al | Liu, et al | PA-2 + SDS | PA-2 + LIRS | Brick house image |

(b)

**Fig. 12** (a) Original hazy images processed with various algorithms and (b) key to figures

**Table 1** Average values for 66 (a) left and (b) right view (FRIDA database) images processed with Zhu, et al (CAP), He, et al, (DCP), PA-2, PA-2 + SDS and PA-2 + LIR

| Algorithm | Qe | BWAR | RAG | MSSIM | PSNR | MAE | MSE |
|---|---|---|---|---|---|---|---|
| CAP | 0.967408 | 0.01001 | 1.617735 | 0.592498 | 11.09096 | 53.86885 | 5415.557 |
| DCP | 0.97796 | 0.00017 | 2.19491 | 0.648064 | 12.56377 | 44.63618 | **3848.073** |
| PA2 | 1.058611 | **8.8E-05** | 2.276011 | 0.611011 | 11.00526 | 63.69751 | 5464.016 |
| PA-2 + SDS | **1.254145** | 1.59308E-05 | **2.632071** | **0.71882** | **12.67592** | **43.88135** | 3910.314 |
| PA-2 + LIR | 1.115049 | 5.86924E-05 | 2.445847 | 0.693955 | 12.59826 | 45.75688 | 3901.96 |

(a)

| Algorithm | Qe | BWAR | RAG | MSSIM | PSNR | MAE | MSE |
|---|---|---|---|---|---|---|---|
| CAP | 0.933719 | 0.00972 | 1.619955 | 0.593161 | 11.0824 | 53.90755 | 5422.409 |
| DCP | 0.954648 | 0.00017 | 2.19394 | 0.648356 | 12.54855 | 44.71968 | **3856.819** |
| PA2 | 1.03751 | 1.13439E-06 | 2.2759 | 0.609253 | 10.95889 | 64.02167 | 5537.211 |
| PA-2 + SDS | **1.214761** | **1.16892E-05** | **2.634608** | **0.71873** | **12.66522** | **43.97349** | 3917.517 |
| PA-2 + LIR | 1.124608 | 1.02588E-05 | 2.447449 | 0.693724 | 12.58288 | 45.87162 | 3914.086 |

(b)

**Table 2** Average values for FIDA images processed with Zhu, et al (CAP), He, et al, (DCP), PA-2, PA-2 + SDS and PA-2 + LIR

| Algorithm | Qe | BWAR | RAG | HDI | CEF |
|---|---|---|---|---|---|
| He (DCP) | 1.707122 | 0.01194 | 1.996111 | 5.140865 | 1.594571 |
| Tan | 1.712774 | 0.0122 | 2.189728 | 4.322899 | 2.543521 |
| Fattal | 1.048314 | 0.0065 | 1.239824 | 5.727238 | 1.132069 |
| Kopf | 1.054118 | 0.00151 | 1.396871 | 3.927113 | 1.06452 |
| Tarel | **2.344905** | **3.5E-07** | **2.685774** | 6.472337 | 1.4694 |
| Tarel2 | 1.626088 | 0.00017 | 1.576127 | N/A | N/A |
| Zhu (CAP) | 1.152199 | 0.001234 | 1.096838 | 2.331844 | 0.558164 |
| PA2 | 1.188773 | 0.010317 | 1.232618 | 1.338438 | 1.020475 |
| PA2+SDS | 1.477977 | 0.004188 | 1.918533 | **0.383366** | 2.490711 |
| PA2+LIRS | 1.400834 | 0.007545 | 1.801544 | 0.632029 | **2.679427** |

**Table 3** Average values for 53 benchmark images processed with Zhu, et al (CAP), He, et al, (DCP), PA-2, PA-2 + SDS and PA-2 + LIRS

| Algorithm | Qe | BWAR | RAG | HDI | CEF |
|---|---|---|---|---|---|
| He (DCP) | 1.216391 | 0.00014 | 1.959838 | 7.211636 | 2.045449 |
| Zhu (CAP) | 1.101555 | 0.00361 | 1.261084 | 5.240882 | 1.514626 |
| PA2 | **1.270947** | **7.31E-07** | 2.371226 | 2.135789 | **3.010938** |
| PA2+SDS | 1.197107 | 3.73474E-06 | **2.572577** | 2.099481 | 2.484864 |
| PA2+LIRS | 1.212536 | 9.7E-06 | 2.536623 | **2.054884** | 2.509167 |

**Table 4** Runtimes for images processed with algorithms by He et al [4], Zhu, et al [59], Ren, et al and PA-2

| Algos \ Images | He, et al [4] $T_H$ (s) (standard/fast) | Zhu, et al [63] [59] $T_Z$ (s) | Ren et al [45] $T_R$ (s) | PA-2/ PA-2+SDS/ PA-2+LIRS $T_{PA}$ (s) | No. of iterations (N) |
|---|---|---|---|---|---|
| *Tianamen* (450×600) | 30.364180/ 1.253494 | 0.991586 | 2.362754 | 3.200073/4.3571/4.2563 | 118 |
| *Cones* (384×465) | 21.205725/ 0.850155 | 0.661314 | 1.651447 | 2.169653/2.7861/2.6889 | 119 |
| *City1* (600×400) | 28.019656/ 1.094910 | 0.875287 | 2.070620 | 3.011386/3.6758/3.6244 | 117 |
| *Canyon* (600×450) | 30.861073/ 1.237655 | 0.972741 | 2.529734 | 3.413434/4.2049/4.1201 | 119 |
| *Canon* (525×600) | 35.503302/ 1.431257 | 1.135376 | 2.890541 | 3.899115/4.8064/4.673 | 118 |
| *Mountain* (400×600) | 27.063493/ 1.129231 | 0.880835 | 2.358143 | 3.074064/3.6307/3.7172 | 119 |
| *Brickhouse* (711×693) | 57.417363/ 2.230871 | 1.667610 | 5.234674 | 5.826943/7.4915/7.6374 | 116 |
| *Pumpkins* (400×600) | 27.23115/ 1.125475 | 0.901815 | 2.253179 | 2.967405/3.9021/3.448 | 120 |
| *Train* (400×600) | 27.443427/ 1.105757 | 0.849072 | 2.075004 | 3.064349/4.4596/4.01 | 118 |
| *Toys* (360×500) | 20.462751/ 0.844945 | 0.657376 | 1.578068 | 2.282819/2.9972/2.7936 | 118 |

**Table 5** Runtimes for images processed with algorithms by He et al [4], Zhu, et al [59], Ren, et al, PA-2, PA-2 + SDS and PA-2 + LIR

| Algorithms \ Images | He (standard/ fast) | Zhu | Ren | PA2 | PA2+SDS | PA2+LIRS | Image dimensions | |
|---|---|---|---|---|---|---|---|---|
| 'Faisal_Mosque_10.jpg' | 3.3421 | 2.9726 | - | 17.0872 | 16.0226 | 15.4859 | 768 | 1024 |
| 'Haze-KL-Jul-2011.jpg' | 0.374 | 0.3614 | - | 1.6932 | 1.3321 | 1.1895 | 240 | 421 |
| 'Haze-induced.jpg' | 0.3951 | 0.3936 | - | 1.4197 | 1.2761 | 1.1271 | 267 | 400 |
| 'HazeK_Lumpur.jpg' | 6.07 | 4.6331 | - | 25.3979 | 26.3608 | 25.0605 | 960 | 1280 |
| 'HazeK_Lumpur2.JPG' | 15.9383 | 12.6068 | - | 81.1902 | 82.9265 | 84.0672 | 1536 | 2048 |
| 'MALAYSIA_HAZE.jpg' | 1.0218 | 0.9632 | -- | 3.4821 | 3.7777 | 3.7621 | 382 | 630 |
| 'Riau_Haze.JPG' | 37.2645 | 31.0729 | - | 179.643 | 192.0879 | 187.1328 | 2247 | 3500 |
| 'SINGAPOREHAZE.jpg' | 7.1098 | 6.0222 | - | 40.6832 | 38.8779 | 38.0073 | 1024 | 1536 |
| 'ST_HAZE18LE.jpg' | 1.1647 | 1.1607 | - | 4.2314 | 4.8418 | 4.5567 | 410 | 722 |
| 'haze-singapore.jpg' | 0.3881 | 0.3847 | - | 1.3173 | 1.5929 | 1.2816 | 281 | 375 |
| 'haze-singapore2.jpg' | 1.3495 | 1.2873 | - | 4.9798 | 5.396 | 5.2063 | 432 | 768 |
| 'haze21.jpg' | 38.9051 | 30.7838 | - | 180.4488 | 188.3506 | 187.284 | 2292 | 3445 |
| 'haze23e.jpg' | 1.707 | 1.1324 | - | 4.4489 | 5.1237 | 4.4685 | 410 | 722 |
| 'haze_2596971b.jpg' | 0.9678 | 0.8913 | - | 3.4788 | 4.1412 | 3.5047 | 387 | 620 |
| 'indonesia-haze.jpg' | 2.5846 | 2.2883 | - | 10.3493 | 11.2843 | 10.6957 | 641 | 950 |
| 'june-17-haze-ugc.png' | 1.3707 | 1.2903 | - | 5.114 | 5.5174 | 5.1336 | 432 | 768 |
| 'st_haze2_0.jpg' | 0.589 | 0.7064 | - | 1.639 | 1.7846 | 1.6837 | 280 | 430 |
| '9.png' | 57.4174/2.0919 | 1.7942 | 5.2347 | 7.5514 | 7.4915 | 7.6374 | 711 | 693 |
| 'IMG_0752.png' | 1.3143 | 1.2718 | | 5.0409 | 5.1733 | 5.2072 | 432 | 768 |
| 'canon.jpg' | 35.5033/1.2287 | 1.2237 | 2.8905 | 4.9874 | 4.8064 | 4.673 | 525 | 600 |
| 'canon3.bmp' | 0.9403 | 0.8735 | - | 3.5412 | 3.6216 | 3.7172 | 600 | 400 |
| 'canyon.bmp' | 30.8611/1.1319 | 1.0435 | 2.5297 | 4.2559 | 4.2049 | 4.1201 | 600 | 450 |
| 'city_1.jpg' | 28.0197/ 0.9305 | 0.8608 | 2.0706 | 3.6774 | 3.6758 | 3.6244 | 600 | 400 |
| 'city_2.jpg' | 0.9292 | 0.9151 | - | 3.588 | 3.6914 | 3.6354 | 400 | 600 |
| 'cones.jpg' | 21.2057/ 0.7196 | 0.6603 | 1.6514 | 2.7228 | 2.7861 | 2.6889 | 384 | 465 |
| 'example-04-haze.png' | 1.6046 | 1.478 | - | 6.086 | 6.3906 | 6.1763 | 768 | 511 |
| 'foggy-morning.jpg' | 0.7104 | 0.6641 | - | 2.5385 | 2.8326 | 2.6969 | 492 | 360 |

| | | | | | | | |
|---|---|---|---|---|---|---|---|
| 'forest.jpg' | 3.3938 | 2.9365 | - | 13.017 | 14.4966 | 15.0649 | 768 | 1024 |
| 'gugong.bmp' | 0.992 | 0.8957 | - | 3.506 | 3.7649 | 3.6629 | 600 | 400 |
| 'haze.jpg' | 0.6794 | 0.6867 | - | 2.5417 | 2.9184 | 2.8355 | 341 | 512 |
| 'hazed.jpg' | 1.1613 | 1.1625 | - | 4.0827 | 4.5786 | 4.198 | 609 | 462 |
| 'hazed3.jpg' | 0.795 | 0.7623 | - | 3.0964 | 3.3758 | 3.0446 | 339 | 598 |
| 'hazed4.jpg' | 1.1778 | 0.9321 | - | 4.6179 | 4.1444 | 3.8017 | 425 | 600 |
| 'hazed5.jpg' | 1.7532 | 1.6374 | - | 7.0024 | 7.1662 | 6.9139 | 756 | 576 |
| 'hazed7.jpg' | 0.648 | 0.5852 | - | 2.9081 | 2.5143 | 2.415 | 458 | 361 |
| 'hazy-kl.jpg' | 7.2913 | 6.7634 | - | 38.749 | 37.7926 | 37.1421 | 1080 | 1628 |
| 'hazy-pictures.jpg' | 8.1555 | 7.4005 | - | 44.2187 | 41.1485 | 41.7144 | 1200 | 1600 |
| 'hazy_buses.jpg' | 2.1928 | 1.9841 | - | 9.1077 | 8.8751 | 9.1441 | 585 | 860 |
| 'hazy_cities.jpg' | 1.3865 | 1.4062 | - | 5.6987 | 5.7986 | 5.8182 | 450 | 800 |
| 'hongkong.bmp' | 1.4214 | 1.3359 | - | 5.6883 | 5.8078 | 5.8033 | 457 | 800 |
| 'image.jpg' | 0.473 | 0.4753 | - | 1.6684 | 1.7516 | 1.7163 | 284 | 456 |
| 'karlskrona3.jpg' | 0.9413 | 1.1008 | - | 3.6914 | 3.642 | 3.7338 | 400 | 600 |
| 'lake.jpg' | 0.581 | 0.5484 | - | 0.7625 | 0.7961 | 2.078 | 321 | 459 |
| 'mountain.jpg' | 27.0635/0.971 | 0.9403 | 2.3581 | 3.6207 | 3.6307 | 3.7172 | 400 | 600 |
| 'ny1.bmp' | 1.4126 | 1.3101 | - | 5.3571 | 5.4747 | 5.4522 | 431 | 800 |
| 'ny_1.jpg' | 1.8306 | 1.638 | - | 7.3589 | 7.7931 | 7.4311 | 768 | 576 |
| 'ny_2.jpg' | 3.3266 | 2.8593 | - | 15.5844 | 17.02 | 15.3891 | 768 | 1024 |
| 'pumpkins.jpg' | 27.2312/ 0.9822 | 0.9137 | 2.2532 | 3.5153 | 3.9021 | 3.448 | 400 | 600 |
| 'screenshot.jpg' | 0.5996 | 0.5519 | - | 2.0616 | 2.2326 | 2.0414 | 309 | 468 |
| 'stadium.jpg' | 1.3556 | 1.2681 | - | 4.9069 | 5.2228 | 4.8464 | 327 | 1000 |
| 'tiananmen1.png' | 30.3641/ 1.1334 | 1.038 | 2.3628 | 4.19 | 4.3571 | 4.2563 | 450 | 600 |
| 'toys.jpg' | 0.7599 | 0.6761 | 1.5781 | 2.7306 | 2.9972 | 2.7936 | 360 | 500 |
| 'train.bmp' | 27.4434/1.1409 | 1.0684 | 2.0750 | 3.9325 | 4.4596 | 4.01 | 400 | 600 |

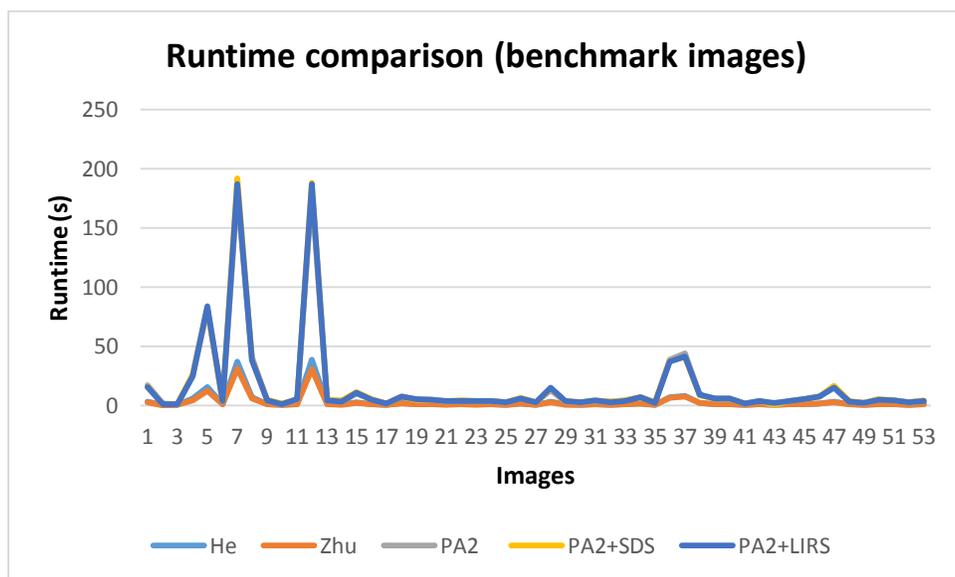

(a)

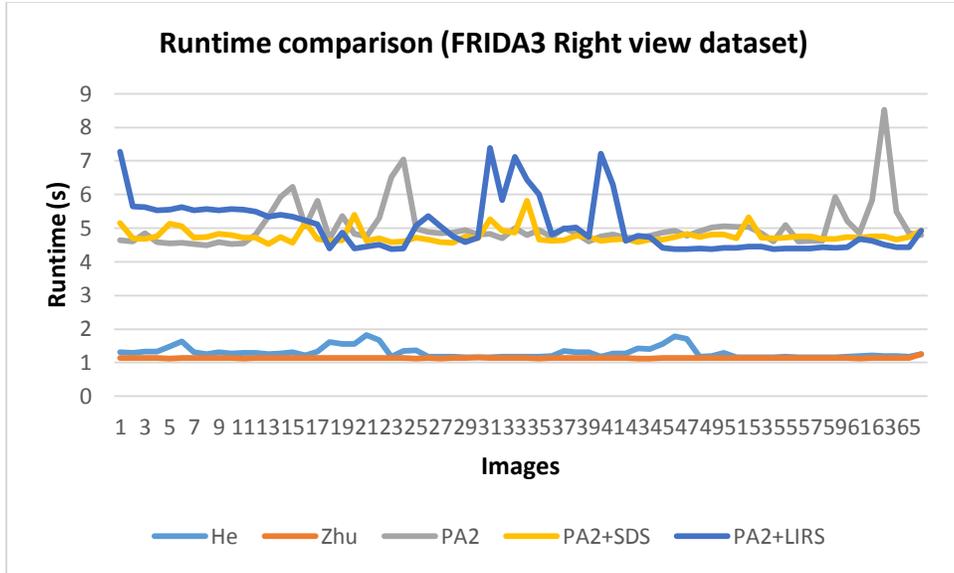

(b)

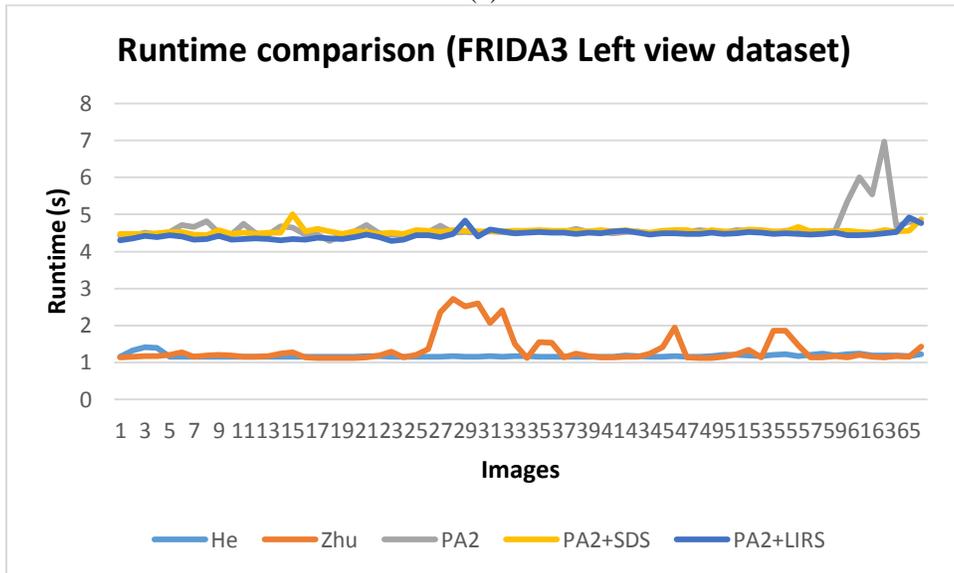

(c)

**Fig. 13** Runtime comparison for (a) benchmark images and FRIDA3 dataset (b) right and (c) left view images processed with He, et al, Zhu, et al [59], PA2 [22], PA2 + SDS and PA2 + LIR

## 5. Conclusion

This report has presented the results of a sky detection and enhancement technique and an alternative log illumination refinement scheme [46] for improvement of results of a previously developed PDE-based hazy image contrast enhancement algorithm [22]. The schemes have been verified using benchmark image datasets based on visual and quantitative evaluation. However, the schemes contribute to the execution time of the system. Future work will be based on reducing the run-time and computational complexity of the algorithms, while improving their effectiveness.